\documentclass[runningheads]{llncs}

\usepackage{eccv}

\definecolor{myblue}{RGB}{58,74,145}
\definecolor{myorange}{RGB}{221,129,5}

\usepackage{array,booktabs,makecell}
\usepackage{xspace}
\usepackage{xcolor}   
\usepackage{multirow} 
\usepackage{graphicx}

\usepackage{eccvabbrv}

\usepackage{graphicx}
\usepackage{booktabs}

\usepackage[accsupp]{axessibility}  

\newcommand{\model}{\textit{ReCamDriving}\xspace}
\newcommand{\dataset}{\textit{ParaDrive}\xspace}

\usepackage{hyperref}

\usepackage{orcidlink}

\begin{document}

\title{ReCamDriving: LiDAR-Free Camera-Controlled Video Synthesis for Novel Trajectories}
\titlerunning{ReCamDriving}


\author{
Li Yaokun\inst{1}\thanks{This work was done during Yaokun Li's internship at ZhuoYu Technology.}
\and
Shuaixian Wang\inst{1,3}
\and
Mantang Guo\inst{2}
\and
Jiehui Huang\inst{4}
\and
Taojun Ding\inst{2}
\and
Mu Hu\inst{4}
\and
Kaixuan Wang\inst{2}
\and
Shaojie Shen\inst{4}
\and
Guang Tan\inst{1}\textsuperscript{\dag}
}

\authorrunning{Y. Li et al.}



\institute{
Sun Yat-sen University \and
ZhuoYu Technology \and
Shenzhen Polytechnic University \and
The Hong Kong University of Science and Technology
}

\begingroup
\renewcommand{\thefootnote}{\dag}
\footnotetext{Corresponding author.}
\endgroup

\maketitle

\begin{abstract}
  Synthesizing multi-pass videos is important for autonomous driving. While current repair-based methods often struggle with out-of-distribution artifacts, camera-controlled methods often produce 3D-inconsistent results due to sparse LiDAR cues. We propose ReCamDriving, a purely vision-based framework that achieves camera-controlled generation by leveraging dense, structurally complete 3DGS renderings as geometric guidance. Specifically, to prevent the model from overfitting to a trivial repair solution when conditioning on 3DGS renderings, we adopt a two-stage progressive training paradigm: the first stage uses camera poses for coarse control, while the second stage incorporates 3DGS renderings for fine-grained viewpoint and geometric guidance. Furthermore, to align training and inference camera transformation patterns, we propose a 3DGS-based cross-trajectory data curation strategy, enabling consistent lateral-trajectory supervision from single-pass videos. Based on this strategy, we construct the ParaDrive dataset, containing approximately 110K parallel-trajectory video pairs. Extensive experiments demonstrate that ReCamDriving achieves state-of-the-art camera controllability and structural consistency. Project website: \href{https://recamdriving.github.io/}{https://recamdriving.github.io/}.
  \keywords{Video generation \and Autonomous driving \and Novel trajectory synthesis}
\end{abstract}    

\vspace{-2mm}
\section{Introduction}
\label{sec:intro}

High-quality multi-pass videos is pivotal for autonomous driving, as it provides diverse viewpoints for tasks such as 3D reconstruction~\cite{streetgaussian,flexdrive} and world-model training~\cite{drivedreamer4d,drivinggpt}. However, collecting real-world multi-pass data is costly, requiring multiple synchronized vehicles to capture videos from different trajectories. Consequently, 3D-consistent, high-fidelity novel-trajectory video synthesis from single-pass ego-trajectories offers an attractive, scalable alternative.

\begin{figure}[!t]
\centering{\includegraphics[width=\linewidth]{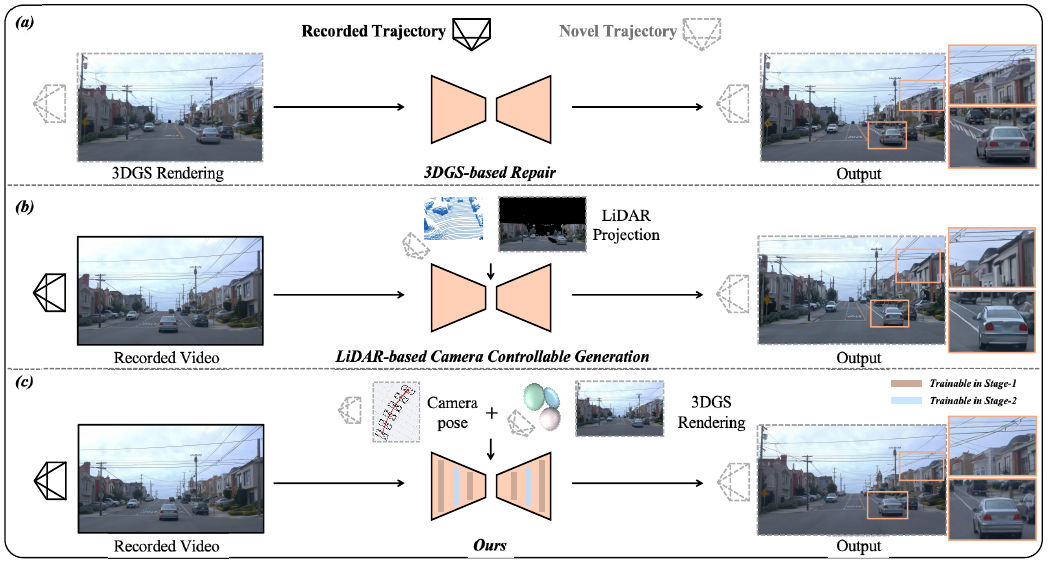}}
\caption{Comparison of novel-trajectory video synthesis. Repair-based methods (e.g., Difix3D+~\cite{difix3d+}) suffer from severe artifacts under large-extrapolation novel viewpoints, while LiDAR-based camera-controlled methods (e.g., StreetCrafter~\cite{StreetCrafter}) exhibit inconsistencies in occluded or distant regions due to incomplete geometric cues. In contrast, ReCamDriving leverages dense scene-structure information of novel-trajectory 3DGS renderings to achieve precise camera control and structurally consistent generation.}
\label{fig:teaser}
\vspace{-1.2em}
\end{figure}

A feasible strategy for novel-trajectory synthesis is the reconstruction-then-repair pipeline~\cite{difix3d+,freesim,gsfixer,3dgs-enhancer}. It first reconstructs scenes using Neural Radiance Fields (NeRF)~\cite{nerf} or 3D Gaussian Splatting (3DGS)~\cite{3dgs}, renders novel trajectories, and then trains diffusion-based repair models to restore rendering artifacts. Although effective at artifact removal, this pipeline often fails under complex rendering artifacts (Fig.~\ref{fig:teaser}a). Its core limitation is that repair models learn local degraded-to-clean mappings based on training-time artifact patterns, whereas highly varying inference-time artifacts often fall out-of-distribution (OOD), leading to failed restoration and 3D inconsistency.

Another paradigm is camera-controlled video generation~\cite{Recammaster,Recapture}, which synthesizes novel-trajectory videos based on source video and camera conditions. Early works~\cite{motionctrl,ac3d,CameraCtrl} simply condition on camera poses, often leading to imprecise viewpoint control. To enhance accuracy, FreeVS~\cite{FreeVS} and StreetCrafter~\cite{StreetCrafter} incorporate novel-trajectory LiDAR projections to provide explicit geometric cues, yet LiDAR projections are sparse and incomplete in background or occluded regions, often resulting in 3D-inconsistent results (Fig.~\ref{fig:teaser}b). Moreover, training camera-transformation models requires ground-truth novel-trajectory supervision, which is unavailable in autonomous driving datasets. Prior works~\cite{FreeVS,StreetCrafter} circumvent this by constructing pseudo training pairs from different segments of the single-pass recorded trajectory (Fig.~\ref{fig:data_curation}a). However, this compromise only captures longitudinal motion patterns, leading to a train-test gap when generalizing to lateral novel trajectory synthesis at inference time.

To address these limitations, we propose \model, a purely vision-based framework for 3D-consistent novel-trajectory video synthesis. Our approach redefines camera-controlled generation by replacing sparse LiDAR projections with novel-trajectory 3DGS renderings as the primary control signal. The key insight is that the structural completeness of 3DGS renderings, despite potential rendering artifacts, is more critical for precise camera control and structurally consistent generation.

However, directly conditioning the model on 3DGS renderings introduces a significant risk of shortcut learning. Since the conditioning renderings and the ground-truth videos share the same viewpoint, the model tends to prioritize local artifact restoration over complex novel-trajectory geometric reasoning, thereby collapsing into a trivial artifact-repair solution. To circumvent this, we propose a two-stage progressive training strategy. We first train the model using only camera poses to establish coarse control capability. Subsequently, we freeze these modules and introduce auxiliary attention layers to incorporate 3DGS renderings. This decoupling simplifies novel-trajectory reasoning by interacting rendering tokens with latent features pre-aligned to the target pose. This encourages the model to treat 3DGS renderings as a structural scaffold for guidance rather than a template for restoration.

Furthermore, to bridge the train-test gap of camera transformation patterns, we introduce a 3DGS-based cross-trajectory data curation strategy (Fig.~\ref{fig:data_curation}). By utilizing novel-trajectory 3DGS renderings as source views and recorded videos as supervision during training, we align the camera-transformation patterns encountered at inference. This strategy enables large-scale lateral-trajectory supervision from monocular videos, allowing us to construct the \dataset dataset, comprising 110K parallel-trajectory video pairs across 1.6K 3DGS scenes from Waymo Open Dataset (WOD)~\cite{waymo} and NuScenes~\cite{nuscenes}. This approach facilitates scalable data construction and has the potential to extend camera-controlled models to web-scale video data without requiring LiDAR annotations. Overall, our main contributions are:
\begin{itemize}
    \item We propose \model, a vision-based framework leveraging 3DGS renderings for precise camera control and structurally consistent novel-trajectory synthesis.
    \item We introduce a novel 3DGS-based cross-trajectory data curation strategy for scalable lateral-trajectory supervision and construct the \dataset dataset with 110K pairs.
    \item Extensive experiments demonstrate that ReCamDriving achieves the state-of-the-art performance in both camera controllability and 3D consistency.
\end{itemize}

\section{Related Work}
\label{sec:related}

\subsection{Diffusion Priors for Repairing 3D Renderings.}  
Recent advances in NeRF~\cite{nerf} and 3DGS~\cite{3dgs} have greatly advanced 3D reconstruction~\cite{mip-nerf,nerf-in-the-wild,mip-nerf-360,pr,mip-splatting,id-nerf-1,bilateral,ie-nerf,alignerf}. However, under limited viewpoints, these methods still produce noticeable artifacts during novel-view synthesis, especially in extrapolated regions~\cite{sparf,id-nerf-2,sparse3d}. To mitigate this issue, recent works explore learning diffusion priors to repair degraded 3D renderings. 3DGS-Enhancer~\cite{3dgs-enhancer} fine-tunes a video diffusion model for rendering restoration, Difix3D+~\cite{difix3d+} enables real-time neural enhancement, and GSFixer~\cite{gsfixer} conditions video restoration on both semantic and geometric cues. Freesim~\cite{freesim} introduces a framework for novel-trajectory restoration through a progressive reconstruction strategy. While these methods can improve the visual fidelity of 3D renderings, they essentially address a restoration problem, focusing on local artifact correction rather than consistent scene-level geometry modeling. Consequently, their performance heavily depends on the training data distribution, and since rendering artifacts vary significantly across scenes and viewpoints, they often fail when facing out-of-distribution degradations.

\subsection{Camera-Controlled Video Generation.} 

Content consistency has long been a central research topic in generative modeling. In recent years, building upon advances in identity and content preservation~\cite{viperid,video_id_1,unityvideo,consistentid}, camera-controlled generation has attracted increasing attention for synthesizing immersive scenes and novel viewpoints~\cite{video-survey,Recapture,Gen3c,3DTrajMaster,Trajectory-attention,zo3t,MotionMaster,CineMaster,Recammaster}. The key challenge lies in how to represent camera motion and inject it into generative models for precise control. Early works~\cite{motionctrl,Recammaster,Direct-a-video,CameraCtrl,ac3d} condition diffusion models on camera extrinsics or Plücker embeddings, enabling controllable generation but often resulting in inconsistent target-view geometry. From a representation-learning perspective~\cite{representation-learning-1,chao1,representation-learning-2,chao2}, such pose-conditioned models primarily capture \emph{statistical correlations} between camera motion and appearance variation rather than \emph{geometric causality}, leading to spatial misalignment and geometric instability.

To address these issues, recent approaches~\cite{viewcrafter,trajectorycrafter} incorporate 3D point-map priors~\cite{dust3r,depthcrafter,mast3r} to inject explicit geometric structure. However, their performance remains limited by the quality of reconstructed point maps in large-scale driving scenes. Alternatively, FreeVS~\cite{FreeVS} and StreetCrafter~\cite{StreetCrafter} leverage LiDAR point clouds for camera control, yet LiDAR data are costly and spatially incomplete, frequently leading to geometrically inconsistent generation results. In contrast, our approach adopts a purely visual formulation for novel-trajectory video generation. By replacing LiDAR with 3DGS renderings as dense structural camera conditions, we achieve fine-grained, geometrically consistent control.
\section{Cross-trajectory Data Curation and ParaDrive Dataset}\label{sec:dataset}

\begin{figure}[!t]
\centering{\includegraphics[width=0.98\linewidth]{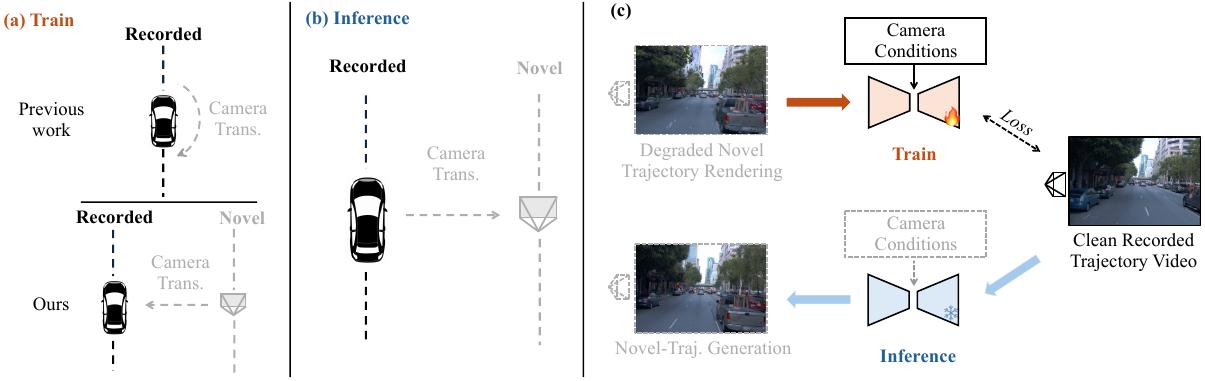}}
\caption{(a–b) Comparison of training and inference camera-transformation patterns. (c) Our training and inference data strategy. (Trans.: Transformation)}
\label{fig:data_curation}
\vspace{-1.3em}
\end{figure}

Training a camera-controlled video regeneration model requires ground-truth videos from novel trajectories for supervision. While general datasets~\cite{Dl3dv-10k,co3d} provide such supervision, autonomous driving datasets typically contain only single-pass trajectories. Prior works~\cite{FreeVS,StreetCrafter} mitigate this limitation by splitting a single trajectory into source and target segments (Fig.~\ref{fig:data_curation}a), but this setting models only longitudinal motion, resulting in degraded performance for lateral-view generation during inference.

To address this challenge, we propose a 3DGS-based cross-trajectory data curation strategy that enables supervised learning of lateral camera transformations. Specifically, given an original trajectory video, we first reconstruct a 3DGS representation of the scene and laterally shift the camera trajectory by a fixed offset (e.g., +3 m) to render a novel trajectory. Since the rendered novel-trajectory video contains artifacts and cannot serve as ground truth, we instead use it as the source input during training, while the clean recorded-trajectory video provides the ground-truth supervision (Fig.~\ref{fig:data_curation}c). At inference, we use the clean recorded video as the source input to generate laterally shifted novel-trajectory views . Although the network is trained with rendered videos as source inputs but tested with clean recorded videos, experiments show that this train–inference source mismatch does not degrade performance. On the contrary, using clean source videos during inference further improves visual quality. Please refer to Sec.~\ref{sec:ablation} for more details.

To construct the training pairs, we reconstruct approximately 1.6K 3DGS scenes from WOD~\cite{waymo} and NuScenes~\cite{nuscenes} using the DriveStudio framework~\cite{drivestudio}. To enhance the model's robustness to novel-trajectory rendering artifacts during inference, we save intermediate 3DGS checkpoints at 100, 500, and 1,000 iterations. These underfitted models are used to generate recorded-trajectory renderings with varying artifact levels, which serve as structural camera conditions during training. Meanwhile, the fully-optimized 30K-iteration models are used to render eight laterally shifted trajectories ($\pm$1 m, $\pm$2 m, $\pm$3 m, and $\pm$4 m) to serve as source videos. Each trajectory contains three 121-frame clips (front, middle, and rear). In total, the 3DGS reconstruction process consumed 8,240 L20 GPU hours and produced approximately 110K dual-trajectory video pairs. We name this dataset \textbf{\dataset}, which will be released to facilitate research on camera-controlled video generation.

\section{\model}\label{sec:method}

\begin{figure*}[!t]
\centering{\includegraphics[width=\linewidth]{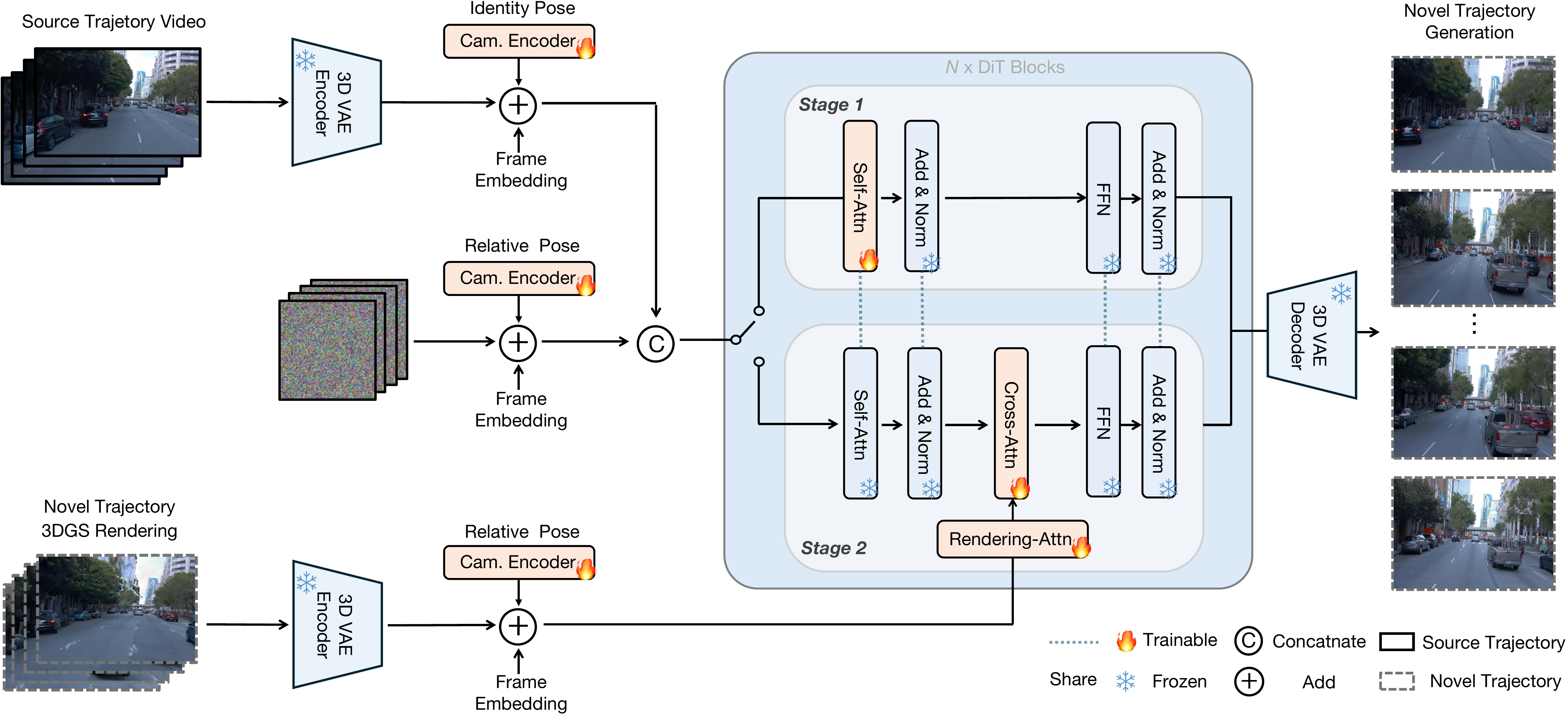}}
\caption{Overview of our framework. We adopt a two-stage training scheme for precise and structurally consistent novel-trajectory video generation. In \emph{Stage~1}, \model trains DiT blocks conditioned on the source trajectory video and relative camera (cam.) pose. When switching to \emph{Stage~2}, the original DiT parameters are frozen, and additional attention modules are introduced to integrate 3DGS renderings for fine-grained view control and structural guidance. Shared modules between stages are connected with blue dashed lines.}
\label{fig:framework}
\vspace{-1.3em}
\end{figure*}

Given a source trajectory video $V_s$, our goal is to synthesize a novel trajectory video $V_t$ with lateral offsets. To this end, we adopt a two-stage training strategy that progressively guides the model to perform viewpoint transformation, as illustrated in Fig.~\ref{fig:framework}. In Stage~1, we aim for the network to understand and learn the physical process of viewpoint transformation by using the relative camera pose $\Delta T = T_{t \leftarrow s} \in SE(3)$ between trajectories, enabling it to warp information from the source video to the target trajectory. In Stage~2, we introduce novel-trajectory 3DGS renderings as fine-grained guidance, freeze the first-stage parameters, and add two additional attention modules to incorporate 3DGS renderings for accurate viewpoint and structural generation.

\subsection{Preliminary: Latent Diffusion Models with Flow Matching}

Latent Diffusion Models (LDMs)~\cite{sd} perform diffusion in a compact latent space learned by a pretrained autoencoder, achieving high-fidelity generation with reduced computational cost. This balance between efficiency and quality has made LDMs the foundation of many state-of-the-art image and video generators~\cite{sd,svd,wan}. Given a video $V \in \mathbb{R}^{F \times 3 \times H \times W}$, a 3D VAE encoder projects it into latents $x \in \mathbb{R}^{f \times c \times h \times w}$ with compressed spatial and temporal dimensions. Diffusion and generation are then performed in this latent space, and the decoded outputs reconstruct the final video frames.

Traditional diffusion models~\cite{ddpm,score-matching,sde-diffusion} formulate generation as a stochastic differential equation (SDE), reversing a predefined noise process but suffering from training instability and slow inference. Flow Matching~\cite{flow-matching,rectified-flow} replaces this stochastic formulation with a deterministic ordinary differential equation (ODE) that learns a time-dependent velocity field to transport samples from noise to data, yielding more stable training and faster sampling. The forward process is typically defined as a straight path between noise $x_0$ and data $x_1$:

\begin{equation}
    x_t = tx_1 + (1 - t)x_0, \quad t \in [0, 1].
\end{equation}
Differentiating with respect to $t$ gives the corresponding velocity field:
\begin{equation}
    v_t = \frac{d x_t}{d t} = x_1 - x_0.
\end{equation}
In the context of camera-controlled video generation, given a camera condition $c_{\text{cam}}$, the training objective of flow matching can be formulated as:
\begin{equation}\label{eq:flow_matching}
    \mathcal{L}_{\text{FM}} 
    = \mathbb{E}_{x_0,\,x_1,\,c_{\text{cam}},\,t \sim U(0,1)} 
    \bigl\| \epsilon_\theta(x_t; c_{\text{cam}}, t) - v_t \bigr\|^2 ,
\end{equation}
where $\epsilon_\theta$ denotes the neural network predicting the velocity field. 

\subsection{Training Stage 1: Relative Pose-Guided Coarse Camera Control}
In this stage, the relative camera pose $\Delta T$ is used as a condition to guide the model in re-generating the source-trajectory video. Instead of using absolute poses, relative poses are easier to specify during inference and are more robust to calibration errors. ReCamMaster~\cite{Recammaster} adopts a similar idea but concatenates the relative pose with both the source latent $x_s$ and the noisy latent $x_t$, which can introduce ambiguity regarding which latent corresponds to the source or the target view.

To resolve this, we separately encode the relative pose $\Delta T$ and the identity pose $T_I = I_4 \in SE(3)$ (representing zero motion) using a camera encoder $\mathcal{E}{_\text{cam}}$, producing $c_r = \mathcal{E}_{\text{cam}}(\Delta T) \in \mathbb{R}^{f \times d}$ and $c_I = \mathcal{E}_{\text{cam}}(T_I) \in \mathbb{R}^{f \times d}$, where $d$ denotes the feature dimension. We further introduce a learnable frame embedding $E_f \in \mathbb{R}^{f \times d}$ that provides frame-wise correspondence cues, enhancing temporal alignment in self-attention operations.

After encoding the source and noisy videos using the 3D VAE encoder, we obtain latent representations $x_s, x_t \in \mathbb{R}^{f \times c \times h \times w}$. A patchify operation unfolds each spatial feature map into a sequence of local tokens, resulting in $x_s, x_t \in \mathbb{R}^{f \times l \times d}$, where $l = h \times w$. The camera and frame embeddings are then added to each latent via broadcasting along the $l$-dimension:
\begin{equation}
    x_i = \mathrm{Cat}(x_t + c_r + E_f,~x_s + c_I + E_f),
\end{equation}
where $\mathrm{Cat}$ denotes concatenation along the frame dimension $f$. The resulting sequence $x_i$ is processed by $N$ layers of Diffusion Transformer (DiT)~\cite{dit} blocks to predict the velocity field. Each DiT block in this stage (see Fig.~\ref{fig:framework}) consists of a self-attention layer, a feed-forward network (FFN), and normalization layers, with only the self-attention parameters unfrozen for adaptation.

We adopt the flow-matching framework to schedule noise levels and optimize the diffusion process. We adopt the flow-matching framework to schedule noise levels and optimize the diffusion process. The training objective follows Eq.~\ref{eq:flow_matching}, where $c_{\text{cam}}$ denotes the relative camera-pose embedding and the target data $x_1$ corresponds to the clean recorded-trajectory videos.

\subsection{Training Stage 2: Fine-grained Camera Control via 3DGS Renderings}
In this stage, to improve camera control accuracy and structural guidance, besides the relative camera pose, we utilize DriveStudio~\cite{drivestudio} to reconstruct 3DGS representations and render novel-trajectory 3DGS renderings as additional conditions. The renderings $V_{gs}$ are encoded by the same 3D VAE encoder and patchified to obtain $x_{gs} \in \mathbb{R}^{f \times l \times d}$. We then augment $x_{gs}$ with the relative camera-pose embedding and frame embedding:
\begin{equation}
    \bar{x}_{gs} = x_{gs} + c_r + E_f.
\end{equation}

To integrate 3DGS features effectively, each DiT block introduces two additional components: a \textit{Rendering Attention} and a \textit{Cross Attention}. The \textit{Rendering Attention} serves as an auxiliary self-attention layer that refines spatio–temporal representations within the 3DGS rendering latent space. It shares the same structure as the self-attention used in stage~1 but is named differently to avoid confusion. The \textit{Cross Attention} then fuses the rendering latent features $\bar{x}_{gs}$ with the diffusion latent $\bar{x}_i = \text{SelfAttn}(x_i)$ to achieve geometric alignment between the generated and target trajectories. The self-attention modules from stage~1 remain frozen to ensure that $\bar{x}_i$ has already undergone coarse camera transformations when interacting with the 3DGS rendering features.

The primary advantage of this two-stage design over a single-stage alternative lies in the reduced complexity of geometric alignment. By the time the diffusion latents interact with the 3DGS rendering features, they have already undergone a coarse viewpoint transformation. This significantly reduces the difficulty for the model to align the geometric features of both inputs to infer novel-trajectory geometry, thereby effectively mitigating the tendency to collapse into a trivial restoration of 3DGS rendering artifacts. Training in this stage also follows Eq.~\ref{eq:flow_matching}, but with the condition $c_{\text{cam}}$ extended to include both the relative pose embedding and the 3DGS rendering latent. Finally, during inference, the 3D VAE decoder reconstructs the novel-trajectory video $V_t$ from the stage~2 latent representation, achieving precise camera control and structural consistency.

\begin{table}[!t]
  \centering
  \caption{Quantitative comparison results on WOD~\cite{waymo}, where bold indicates the best performance, and underline denotes the second best.}
  \label{tab:main_comparison}
  \scriptsize
  \setlength{\tabcolsep}{1.4pt}
  \renewcommand{\arraystretch}{1.25}
  \begin{tabular}{
      l !{\vrule width .5pt}
      *{5}{c} !{\vrule width .5pt}
      *{5}{c}}
    \specialrule{1.2pt}{0pt}{2pt}

    \multicolumn{1}{c}{} &
    \multicolumn{5}{|c|}{\textbf{Lateral Offset \boldmath$\pm$1m}} &
    \multicolumn{5}{c}{\textbf{Lateral Offset \boldmath$\pm$2m}} \\
    \cmidrule(lr){2-6} \cmidrule(lr){7-11}

    \textbf{Method} &
    \multicolumn{2}{c}{\textbf{Visual Quality}} &
    \multicolumn{3}{c|}{\textbf{View Consistency}} &
    \multicolumn{2}{c}{\textbf{Visual Quality}} &
    \multicolumn{3}{c}{\textbf{View Consistency}} \\
    \cmidrule(lr){2-3} \cmidrule(lr){4-6}
    \cmidrule(lr){7-8} \cmidrule(lr){9-11}

    & IQ$\uparrow$ & TCE$\downarrow$ & FID$\downarrow$ &  FVD$\downarrow$ & CLIP-V$\uparrow$
    & IQ$\uparrow$ & TCE$\downarrow$  & FID$\downarrow$ &  FVD$\downarrow$ & CLIP-V$\uparrow$ \\
    \midrule
    DriveStudio & 52.13 & 7.93 & 83.32 & 25.37 & 94.78& 47.32 & \underline{8.36}  &104.24 &39.79 & 94.23 \\
    Difix3D+ & \underline{64.24} & \underline{7.81} &56.35 &27.80 & 95.32 & 63.11 & 9.39 &57.73 &31.88 & 92.85 \\
    FreeVS & 62.74 & 11.27 & 63.06 & 37.06 & 88.99 &  60.16 & 12.30 & 67.87 & 43.59 & 88.41 \\
    StreetCrafter & 63.57 & 9.72 & \underline{28.18} & \underline{20.51} & \underline{96.01} & \underline{63.78} & 10.61 & \underline{46.78} & \underline{22.81} & \underline{94.74}\\
    \midrule
    \textbf{Ours} & \textbf{65.18} & \textbf{3.38}& \textbf{13.76} & \textbf{13.27} & \textbf{97.96} & \textbf{65.34}  &  \textbf{3.69} & \textbf{25.01} & \textbf{14.08} & \textbf{97.18} \\
    \specialrule{1.2pt}{4pt}{2pt}

    \multicolumn{1}{c}{\textbf{Method}} &
    \multicolumn{5}{|c|}{\textbf{Lateral Offset \boldmath$\pm$3m}} &
    \multicolumn{5}{c}{\textbf{Lateral Offset \boldmath$\pm$4m}} \\

    \midrule
    DriveStudio  & 43.83 & \underline{9.06}  &116.12 &63.21& 90.37 & 41.47 & \underline{9.49}  &144.05 &72.50 & 88.76 \\
    Difix3D+  & \underline{60.88} &  9.79  & 66.39 & 45.23 & 91.76 & 58.81 &  10.49  & 78.08 & 65.37 & 90.12 \\
    FreeVS   & 57.71 & 13.32  & 84.87 & 55.76 & 86.33 & 56.15& 14.35 & 107.04 & 58.39 & 85.17 \\
    StreetCrafter & 59.34 & 11.49  & \underline{50.75} &\underline{30.26} & \underline{92.13} &\underline{59.89} & 12.38  & \underline{68.73}& \underline{36.67}& \underline{91.17} \\
    \midrule
    \textbf{Ours}  & \textbf{62.68} & \textbf{3.99}  & \textbf{28.38} & \textbf{22.59} & \textbf{96.50} & \textbf{61.32} &  \textbf{4.30} & \textbf{32.36} & \textbf{26.76} & \textbf{94.91} \\
    
    \specialrule{1.2pt}{0pt}{2pt}

  \end{tabular}
  \vspace{-1.1em}
\end{table}

\begin{table*}[!t]
  \centering
  \caption{Quantitative comparison of camera accuracy on WOD~\cite{waymo}. RErr.: Rotation Error; TErr.: Translation Error.}
  \label{tab:camera_accuracy}
  \footnotesize
  \setlength{\tabcolsep}{3.3pt} 
  \renewcommand{\arraystretch}{1}
  \begin{tabular}{l cccc cccc}
    \toprule
    \textbf{Method} & \multicolumn{2}{c}{\textbf{Offset $\pm$1m}} & \multicolumn{2}{c}{\textbf{Offset $\pm$2m}} & \multicolumn{2}{c}{\textbf{Offset $\pm$3m}} & \multicolumn{2}{c}{\textbf{Offset $\pm$4m}} \\
    \cmidrule(lr){2-3} \cmidrule(lr){4-5} \cmidrule(lr){6-7} \cmidrule(lr){8-9}
    & RErr.$\downarrow$ & TErr.$\downarrow$ & RErr.$\downarrow$ & TErr.$\downarrow$ & RErr.$\downarrow$ & TErr.$\downarrow$ & RErr.$\downarrow$ & TErr.$\downarrow$ \\
    \midrule
    
    Difix3D+      & \underline{1.36} & \underline{2.42} & \underline{1.64} & \underline{2.66} & 2.01 & \underline{2.97} & \underline{2.68} & 3.12 \\
    FreeVS        & 1.71 & 2.88 & 2.12 & 2.93 & 3.17 & 3.78 & 3.02 & 3.39 \\
    StreetCrafter & 1.52 & 2.53 & 1.79 & 2.77 & \underline{1.91} & 3.13 & 2.87 & \underline{3.03} \\
    \midrule
    \textbf{Ours} & \textbf{1.32} & \textbf{2.37} & \textbf{1.45} & \textbf{2.43} & \textbf{1.63} & \textbf{2.65} & \textbf{1.57} & \textbf{2.73} \\
    \bottomrule
  \end{tabular}
  \vspace{-1.1em}
\end{table*}

\section{Experiments}
\subsection{Experimental setup}
\textbf{{Implementation.}} We train \model on the proposed \dataset dataset in two stages. Each stage is conducted on 64 NVIDIA A100 GPUs for 6,000 steps, with a batch size of 1 and a learning rate of 1e-4. The total training takes approximately 3.5 days. The training resolution is set to 480 $\times$ 832, and each video consists of 121 frames. We initialize our model from the Wan2.1 text-to-video foundation model~\cite{wan}, with the text prompt set to an empty string by default. Specifically, the 3D VAE, rendering attention, cross attention, and FFN modules are initialized from Wan2.1, while the camera encoder and self-attention layers are initialized from ReCamMaster~\cite{Recammaster}.

\textbf{Evaluation Dataset and Metrics.} We select 20 scenes each from WOD~\cite{waymo} and NuScenes~\cite{nuscenes} for validation. We evaluate our method from three perspectives: (1) \emph{Visual Quality}: We use Imaging Quality (IQ)~\cite{Vbench} to assess fidelity. For temporal consistency, we follow~\cite{worldscore} to measure the Temporal Consistency Error (TCE, lower is better) by computing the Average End-Point Error between consecutive frames via optical flow~\cite{SEA-RAFT}. Please refer to the supplementary material for further details. (2) \emph{Camera Accuracy}: Following CamCloneMaster~\cite{camclonemaster}, we adopt the state-of-the-art (SOTA) camera estimation method MegaSaM~\cite{megasam} to recover camera parameters from the generated videos, and evaluate accuracy using rotation and translation error. (3) \emph{View Consistency}: We compute the Fréchet Video Distance (FVD)~\cite{fvd} and Fréchet Image Distance (FID)~\cite{fid} between the source and generated video distributions, and further compute the frame-wise CLIP similarity (CLIP-V) to assess cross-view semantic consistency.

\begin{figure*}[!t]
\setlength{\abovecaptionskip}{5pt}
\centering{\includegraphics[width=\linewidth]{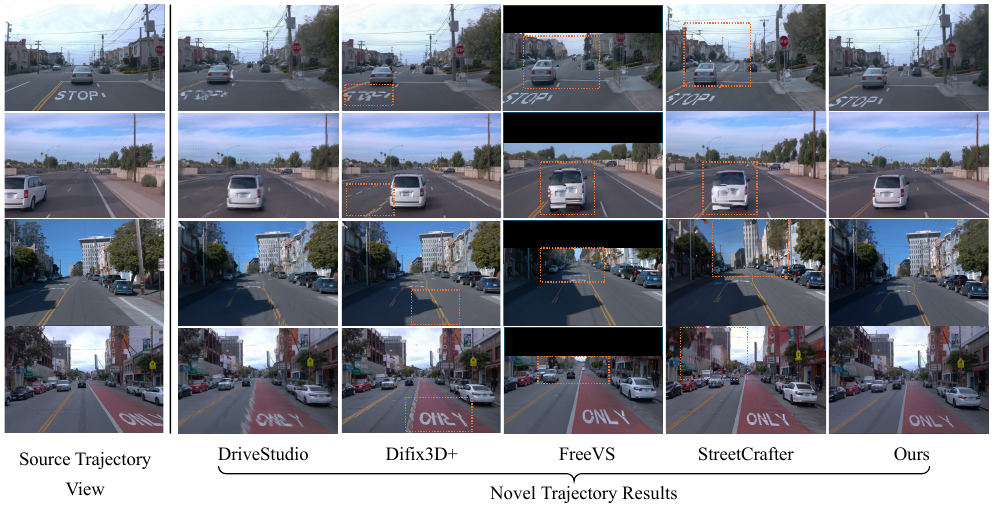}}

\caption{Qualitative comparisons on WOD~\cite{waymo}. Our method and Difix3D+~\cite{difix3d+} use renderings of DriveStudio~\cite{drivestudio} for camera control and restoration, respectively. Note that the officially released FreeVS~\cite{FreeVS} model is trained and tested on a cropped resolution that excludes sky regions to reduce computation and avoid LiDAR-sparse areas.
}
\label{fig:main_results}
\vspace{-0.5em}
\end{figure*}

\begin{figure}[!t]
\setlength{\abovecaptionskip}{5pt}
\centering{\includegraphics[width=0.84\linewidth]{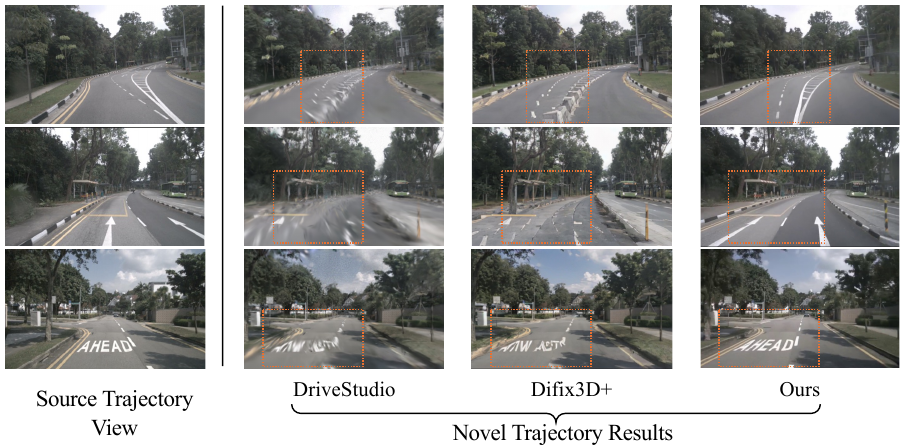}}
\caption{Qualitative comparison results on NuScenes~\cite{nuscenes}.}
\label{fig:main_nuscenes}
\vspace{-1.0em}
\end{figure}

\textbf{Baselines.} We compare \model against state-of-the-art novel trajectory synthesis methods~\cite{drivestudio,difix3d+,FreeVS,StreetCrafter}.
DriveStudio~\cite{drivestudio} reconstructs 3DGS using a dynamic neural scene graph for novel-trajectory rendering. Difix3D+~\cite{difix3d+} follows the reconstruction–repair paradigm, using a single-step diffusion model to restore 3DGS renderings. In addition, FreeVS~\cite{FreeVS} and StreetCrafter~\cite{StreetCrafter} adopt a camera-controll video generation approach, where the target trajectory is conditioned on colored LiDAR point projections to synthesize novel-trajectory videos.

\subsection{Comparison Results} 

\textbf{Qualitative Results.} 
FreeVS~\cite{FreeVS} and StreetCrafter~\cite{StreetCrafter} provide pretrained weights only on WOD. For a fair comparison, we train a version of our model on the WOD subset of \dataset, with qualitative results shown in Fig.~\ref{fig:main_results}.  We further compare our full model trained on the entire \dataset with Difix3D+~\cite{difix3d+} dataset and DriveStudio~\cite{drivestudio} on NuScenes~\cite{nuscenes} (Fig.~\ref{fig:main_nuscenes}). 

\model consistently outperforms all baselines. Difix3D+ often produces view-inconsistent corrections on fine structures near the ego vehicle (e.g., lane markings, road text), whereas \model maintains structural continuity. FreeVS and StreetCrafter exhibit severe inconsistencies under large viewpoint shifts due to sparse LiDAR conditioning. As shown in Fig.~\ref{fig:main_results}, StreetCrafter fails to reconstruct complete vehicle geometry in the first scene, while both baselines yield blurred or 3D-inconsistent distant backgrounds caused by occluded or sparsely sampled LiDAR regions. In contrast, \model leverages dense structural cues from 3DGS renderings for robust view and geometric guidance, enabling more reliable generation of distant content and occluded objects.

\textbf{Quantitative Results.}
Quantitative comparisons on WOD and NuScenes are reported in Table~\ref{tab:main_comparison}, Table~\ref{tab:camera_accuracy} and Table~\ref{tab:main_comparison_avg}. Since DriveStudio renders results using ground-truth poses, we do not evaluate its camera accuracy. As shown, \model consistently outperforms all baselines across all metrics. Notably, as the lateral offset increases, the view consistency of FreeVS~\cite{FreeVS}, StreetCrafter~\cite{StreetCrafter}, and Difix3D+~\cite{difix3d+} degrades sharply, whereas \model remains stable, demonstrating superior robustness under large viewpoint shifts.

\begin{table}[h]
  \vspace{-0.6em}
  \centering
  \caption{Average quantitative results of novel-trajectory generation on NuScenes~\cite{nuscenes}.}
  \label{tab:main_comparison_avg}
  \footnotesize
  \setlength{\tabcolsep}{6pt}
  \renewcommand{\arraystretch}{1.1}

  \begin{tabular}{
      l !{\vrule width .5pt}
      *{2}{c} !{\vrule width .5pt}
      *{3}{c}}
    \specialrule{1.2pt}{0pt}{2pt}

    \multirow{2}{*}{\textbf{Method}} &
    \multicolumn{2}{c|}{\textbf{Visual Quality}} &
    \multicolumn{3}{c}{\textbf{View Consistency}} \\
    \cmidrule(lr){2-3} \cmidrule(lr){4-6}
    & IQ$\uparrow$ & TCE$\downarrow$ & FID$\downarrow$ & FVD$\downarrow$ & CLIP-V$\uparrow$ \\
    \midrule

    DriveStudio & 45.39 & \underline{8.63} & 103.25 & 52.93 & 89.19 \\
    Difix3D+    & \underline{61.76} & 9.14 & \underline{65.14} & \underline{41.37} & \underline{90.48} \\
    \midrule
    \textbf{Ours} & \textbf{62.38} & \textbf{3.79} & \textbf{25.68} & \textbf{18.98} & \textbf{96.14} \\

    \specialrule{1.2pt}{0pt}{2pt}
  \end{tabular}
  \vspace{-2.3em}
\end{table}

\subsection{Ablation Studies}\label{sec:ablation}
\begin{figure}[h]

\setlength{\abovecaptionskip}{7pt}
\centering{\includegraphics[width=\linewidth]{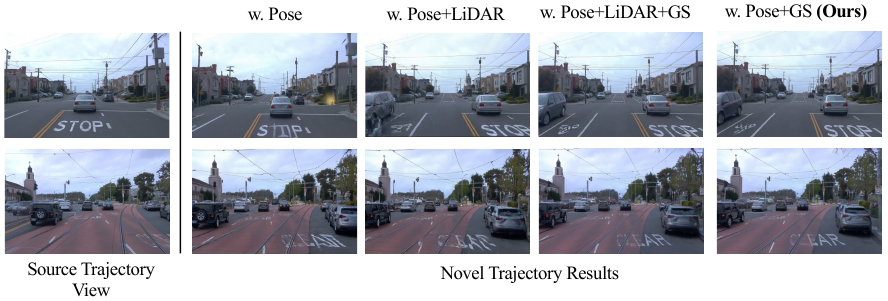}}
\caption{Qualitative ablation of camera conditions on WOD~\cite{waymo}.}
\label{fig:ablation_gs}
\vspace{-0.2em}
\end{figure}

\textbf{Effectiveness of 3DGS Rendering Condition.}
In Stage~2, 3DGS renderings of the target trajectory are used as conditions for fine-grained camera control. We ablate it with three variants: (1) without Stage 2 (\textit{Pose} only); (2) Stage 2 with LiDAR projection (\textit{Pose + LiDAR}); (3) Stage 2 with both LiDAR projection and 3DGS renderings (\textit{Pose + LiDAR + GS}). LiDAR projections are generated following StreetCrafter~\cite{StreetCrafter} and injected into the Rendering Attention branch. Qualitative and quantitative results are presented in Fig.~\ref{fig:ablation_gs} and Table~\ref{tab:ablation 3dgs}. The results demonstrate that using only camera poses often leads to inaccurate camera control. While incorporating LiDAR improves pose accuracy, it still suffers from geometric inconsistencies under large offsets due to the sparsity of LiDAR data. In contrast, our 3DGS-conditioned model effectively mitigates these issues. Furthermore, introducing LiDAR conditions on top of our method does not yield significant improvements in overall metrics, and it incurs high costs for LiDAR data acquisition.

\begin{table}[h]
  \centering
  \caption{Quantitative ablation of camera conditions on WOD~\cite{waymo}.}
  \normalsize
  \label{tab:ablation 3dgs}
  \footnotesize
  \setlength{\tabcolsep}{5.0pt}
  \renewcommand{\arraystretch}{1.05}

  \begin{tabular}{lccccc}
    \specialrule{1.2pt}{0pt}{2pt}
    \textbf{Camera Condition} &
    IQ$\uparrow$& FID$\downarrow$ & FVD$\downarrow$ & RErr.$\downarrow$ & TErr.$\downarrow$   \\
    \midrule
    Pose             & 60.13 & 34.86 & 32.31 & 3.01 & 4.23 \\
    Pose + LiDAR   & 61.32 & 31.23 & 27.78 & 1.53 & 2.69 \\
    Pose + LiDAR + GS & \underline{63.42} & \textbf{24.75} & \underline{19.27} & \textbf{1.41} & \textbf{2.47} \\
    \midrule
    Pose + GS (Ours) & \textbf{63.63} & \underline{24.88} & \textbf{19.18} & \underline{1.49} & \underline{2.55} \\
    \specialrule{1.2pt}{0pt}{2pt}
  \end{tabular}
  \vspace{-1em}
\end{table}

\begin{figure}[h]
\vspace{-0.8em}
\setlength{\abovecaptionskip}{5pt}
\centering{\includegraphics[width=0.84\linewidth]{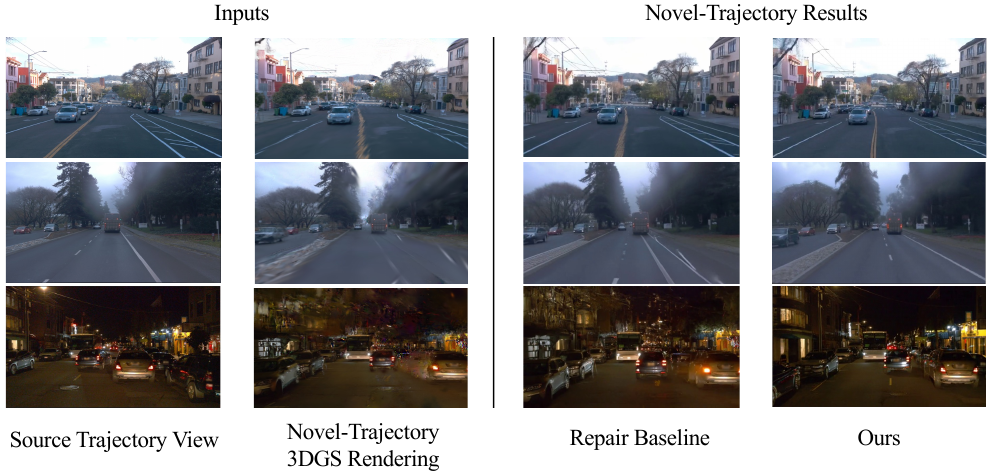}}
\caption{Qualitative ablation of training paradigms on WOD~\cite{waymo}. Our method utilizes the two types of information shown on the left as input. The repair baseline uses novel-trajectory 3DGS renderings as input.}
\label{fig:ablation_repair}
\vspace{-0.8em}
\end{figure}

\textbf{Comparison with the repair baseline.}
We further compare our adopted camera-controlled video generation paradigm with a repair-based baseline. This baseline shares an identical architecture with our framework (Fig.~\ref{fig:framework}) but excludes the Rendering Attention branch and camera pose injection. Specifically, it takes rendered videos as input and produces clean videos as output. For training, we construct pairs on recorded trajectories using blurred 3DGS renderings as inputs and clean videos as supervision. Both methods are evaluated on WOD~\cite{waymo}, with results in Table~\ref{tab:ablation_rerendering} and Fig.~\ref{fig:ablation_repair}. As shown, the repair baseline often fails to correct artifacts, as these artifacts are not well covered by its training distribution.

\begin{table}[h]
  \centering
  \caption{Qualitative ablation of training paradigms on WOD~\cite{waymo}.}
  \label{tab:ablation_rerendering}
  \footnotesize
  \setlength{\tabcolsep}{6pt}
  \renewcommand{\arraystretch}{1.1}

  \begin{tabular}{
      lccccc}
    \specialrule{1.2pt}{0pt}{2pt}

    \textbf{Method} & IQ$\uparrow$ & TCE$\downarrow$ & FID$\downarrow$ & FVD$\downarrow$ & CLIP-V$\uparrow$ \\
    \midrule

    Repair Baseline  & 62.16 & 4.57  & 40.87 & 31.44 & 91.32 \\
    Ours & \textbf{63.63} & \textbf{3.84}  & \textbf{24.88} & \textbf{19.18} & \textbf{96.64} \\
    \specialrule{1.2pt}{0pt}{2pt}
  \end{tabular}
  \vspace{-0.5em}
\end{table}

\begin{figure}[t]

\setlength{\abovecaptionskip}{5pt}
\centering{\includegraphics[width=0.74\linewidth]{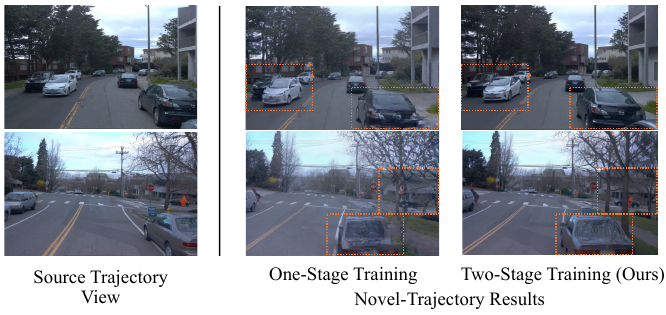}}
\caption{Qualitative ablation on our two-stage training strategy.}
\label{fig:ablation_strategies}
\vspace{-0.6em}
\end{figure}

\textbf{Ablation on the Training Strategies.}
We employ a two-stage strategy to ensure that 3DGS renderings enhance camera control rather than artifact repair. To evaluate its effectiveness, we compare it with a single-stage scheme where all modules are trained jointly without freezing on the full \dataset dataset. Quantitative and qualitative results are shown in Table~\ref{tab:ablation_strategies} and Fig.~\ref{fig:ablation_strategies}. The single-stage model tends to produce artifacts and exhibits degraded visual performance similar to repair-based methods, whereas the two-stage strategy yields clearer and more 3D-consistent results, confirming its effectiveness.

\textbf{Ablation on Cross-Trajectory Data Curation.}
We propose a cross-trajectory data curation strategy to align camera-transformation modes between training and inference. To evaluate its effectiveness, we train a baseline using longitudinal transformations following FreeVS~\cite{FreeVS}, where the first and last 121 frames of each recorded trajectory are used as source and supervision, respectively. Table~\ref{tab:ablation_transfer_mode} presents the comparison results, showing that our cross-trajectory strategy significantly improves both camera accuracy and view consistency for novel-trajectory prediction, confirming its effectiveness.

\begin{figure}[!t]
\centering{\includegraphics[width=0.8\linewidth]{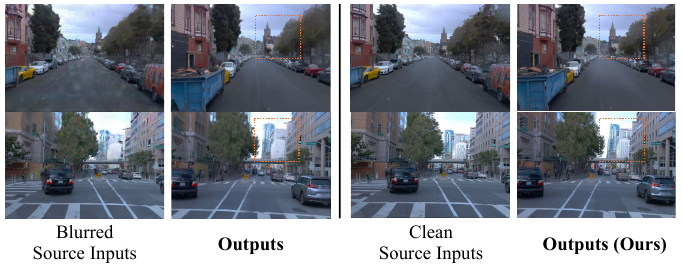}}
\caption{Ablation on different source inputs at inference.}
\label{fig:ablation_train_test_gap}
\vspace{-0.6em}
\end{figure}

\textbf{Impact of Train–Test Source-Video Mismatch.}
During training, we use blurred 3DGS renderings as source videos, while at test time, the source is the clean recorded videos. To examine the impact of this mismatch, we compare inference using either clean recorded videos or their degraded 3DGS-rendered counterparts as sources. As shown in Fig.~\ref{fig:ablation_train_test_gap}, clean sources produce noticeably sharper backgrounds than blurred ones, indicating that clean source videos at test time do not cause degradation but instead improve visual quality. Furthermore, the high sensitivity of outputs to source inputs confirms that our model performs camera-control video generation rather than merely restoring 3DGS rendering conditions, further distinguishing it from repair-based methods.

\begin{table}[!t]
  \centering
  \caption{Quantitative ablation on our two-stage training strategy on WOD~\cite{waymo} and NuScenes~\cite{nuscenes}.}
  \label{tab:ablation_strategies}
  \footnotesize
  \setlength{\tabcolsep}{9pt}  
  \renewcommand{\arraystretch}{1.05}
  
  \begin{tabular}{lcccc}
    \specialrule{1.2pt}{0pt}{2pt}
    \textbf{Training strategy} &
    IQ$\uparrow$ & FID$\downarrow$ & FVD$\downarrow$ & CLIP-V$\uparrow$ \\
    \midrule

    One-stage   & 59.97 & 32.64 & 25.16 & 94.78 \\
    Two-stage (Ours)   & \textbf{63.42} & \textbf{25.13} & \textbf{18.32} & \textbf{96.32} \\
    \specialrule{1.2pt}{0pt}{2pt}
  \end{tabular}

  \vspace{-0.3em}
\end{table}

\begin{table}[!t]
  \centering
  \caption{Ablation of our data curation strategy on WOD~\cite{waymo}.}
  \label{tab:ablation_transfer_mode}
  \footnotesize
  \setlength{\tabcolsep}{4.5pt}
  \renewcommand{\arraystretch}{1.1}
  \begin{tabular}{lcccc}
    \specialrule{1.2pt}{0pt}{2pt}
    \textbf{Camera Transformation} & RErr.$\downarrow$ & TErr.$\downarrow$ & FID$\downarrow$ & FVD$\downarrow$ \\
    \midrule

    Longitudinal   & 1.97 & 3.02 & 34.17 & 28.54 \\
    Lateral (Ours)   & \textbf{1.49} & \textbf{2.55} & \textbf{24.88} & \textbf{19.18} \\
    \specialrule{1.2pt}{0pt}{2pt}
  \end{tabular}
  \vspace{-0.3em}
\end{table}

\vspace{-0.6em}
\section{Discussion and Limitation}
\vspace{-0.3em}
While our method achieves superior performance and eliminates expensive LiDAR acquisition costs, it requires an offline 3DGS pre-processing step. However, this reconstruction is a one-time overhead that supports multiple trajectory generations. Importantly, our framework remains agnostic to the specific reconstruction methodology, meaning that as feed-forward 3D reconstruction~\cite{dggt} techniques continue to evolve, the pre-processing cost can be reduced from hours to seconds. Beyond this overhead, the primary significance of our paradigm lies in offering a scalable pathway to leverage vast, internet-scale datasets that lack expensive LiDAR annotations. 

\vspace{-0.6em}
\section{Conclusion}
\vspace{-0.3em}
In this work, we introduced \model, a pure vision-based framework for controllable novel-trajectory video generation. By leveraging 3DGS renderings as structural camera conditions and adopting a two-stage training paradigm, our model achieves precise camera control and consistent geometry. We also construct the large-scale \dataset dataset using the proposed cross-trajectory data curation strategy, which enables scalable lateral-trajectory supervision from single-pass driving videos. Extensive experiments show that \model achieves higher camera accuracy and visual fidelity than existing baselines.


\section*{Acknowledgements}
The study was supported by the Shenzhen Basic Research Fund under grant JCYJ20241202130025030.

%
%
\bibliographystyle{splncs04}
\bibliography{main}

\clearpage

\section{More Details of Data Construction}
We train 3D Gaussian Splatting (3DGS) on approximately 1.6K scenes from the official Waymo Open Dataset (WOD)~\cite{waymo} \textit{v1.4.3-train} and NuScenes~\cite{nuscenes} \textit{v1.0-train}. Leveraging the DriveStudio framework, we reconstruct each scene over its full temporal duration. Each 3DGS model is trained for 30,000 iterations to render videos along lateral offset trajectories, which serve as the source conditioning inputs during our model's training phase.

\begin{figure}[h]
\centering{\includegraphics[width=0.95\linewidth]{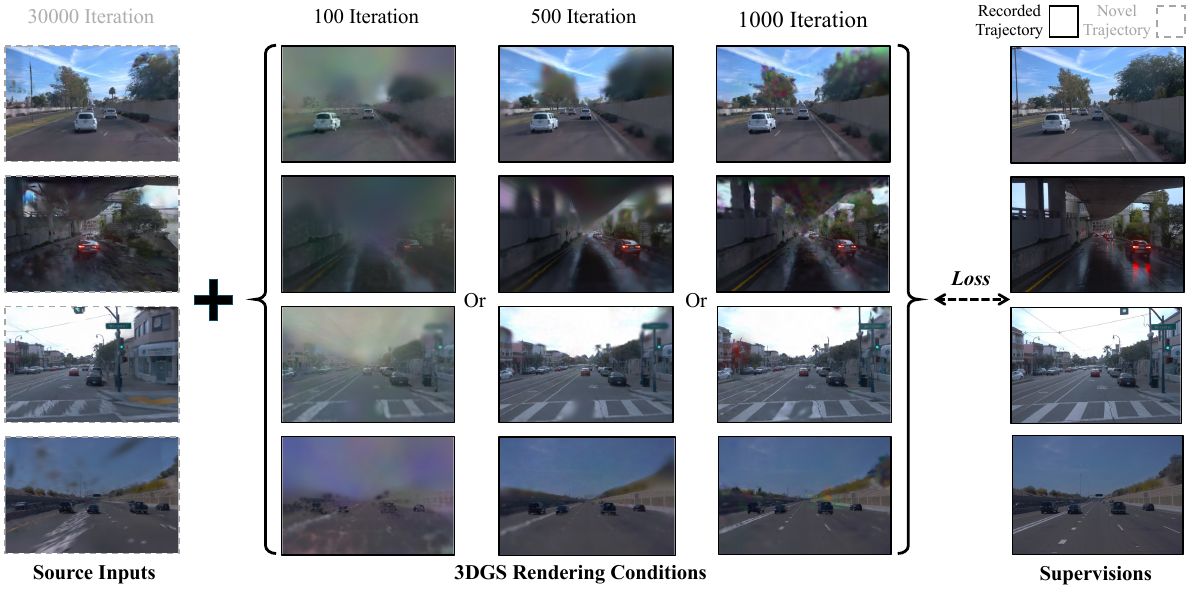}}
\caption{Visualization of training pairs. During training, we use the 3DGS renderings of novel trajectories at 30,000 iterations as the source inputs, the 3DGS renderings of the original trajectory at 100, 500, or 1,000 iterations as the structural rendering conditions, and the clean recorded videos of the original trajectory as the supervision.}
\label{fig:train_pairs}
\end{figure}

During the optimization of each 3DGS scene, we preserve models from multiple intermediate sub-optimal stages (e.g., 100, 500, and 1,000 iterations) to render the recorded trajectories. By utilizing these degraded renderings as training conditions, we aim to closely align the training distribution with the complex degradation patterns encountered during novel-trajectory synthesis at inference time. To ensure these artifacts are representative of real-world reconstruction challenges, we deliberately avoid specialized refinements during 3DGS optimization. For instance, we do not incorporate the SMPL model~\cite{smpl} for articulated human body modeling and maintain the default hyperparameters of the DriveStudio framework. Visualizations of some training pairs are shown in Fig.~\ref{fig:train_pairs}.

\section{Evaluation Dataset}
We evaluate our method on 20 scenes selected respectively from WOD and NuScenes, whose sequence names are listed in Table~\ref{tab:wod_seq} and Table~\ref{tab:nusc_seq}, respectively. For each scene, we use the recorded trajectory video as the source trajectory input and the novel-trajectory 3DGS renderings as the camera-control conditions. Specifically, following the same training setup as in DriveStudio, we train a 3DGS representation for each scene for 30,000 iterations, and then render eight laterally offset novel trajectories ($\pm$1 m, $\pm$2 m, $\pm$3 m, and $\pm$4 m) to serve as the structural camera conditions.

\begin{table}[h]
\centering
\scriptsize
\renewcommand{\arraystretch}{1.25}
\setlength{\tabcolsep}{8.4pt}
\begin{tabular}{c c}
\toprule
\textbf{Dataset} & \textbf{File Names} \\
\midrule
\multirow{20}{*}{WOD}
& \texttt{segment-10444454289801298640\_4360\_000\_4380\_000\_with\_camera\_labels.tfrecord} \\
& \texttt{segment-10498013744573185290\_1240\_000\_1260\_000\_with\_camera\_labels.tfrecord} \\
& \texttt{segment-10588771936253546636\_2300\_000\_2320\_000\_with\_camera\_labels.tfrecord} \\
& \texttt{segment-10625026498155904401\_200\_000\_220\_000\_with\_camera\_labels.tfrecord} \\
& \texttt{segment-10963653239323173269\_1924\_000\_1944\_000\_with\_camera\_labels.tfrecord} \\
& \texttt{segment-11017034898130016754\_697\_830\_717\_830\_with\_camera\_labels.tfrecord} \\
& \texttt{segment-11846396154240966170\_3540\_000\_3560\_000\_with\_camera\_labels.tfrecord} \\
& \texttt{segment-1191788760630624072\_3880\_000\_3900\_000\_with\_camera\_labels.tfrecord} \\
& \texttt{segment-11928449532664718059\_1200\_000\_1220\_000\_with\_camera\_labels.tfrecord} \\
& \texttt{segment-12161824480686739258\_1813\_380\_1833\_380\_with\_camera\_labels.tfrecord} \\
& \texttt{segment-16801666784196221098\_2480\_000\_2500\_000\_with\_camera\_labels.tfrecord} \\
& \texttt{segment-18111897798871103675\_320\_000\_340\_000\_with\_camera\_labels.tfrecord} \\
& \texttt{segment-6242822583398487496\_73\_000\_93\_000\_with\_camera\_labels.tfrecord} \\
& \texttt{segment-1921439581405198744\_1354\_000\_1374\_000\_with\_camera\_labels.tfrecord} \\
& \texttt{segment-2323851946122476774\_7240\_000\_7260\_000\_with\_camera\_labels.tfrecord} \\
& \texttt{segment-1999080374382764042\_7094\_100\_7114\_100\_with\_camera\_labels.tfrecord} \\
& \texttt{segment-4898453812993984151\_199\_000\_219\_000\_with\_camera\_labels.tfrecord} \\
& \texttt{segment-4266984864799709257\_720\_000\_740\_000\_with\_camera\_labels.tfrecord} \\
& \texttt{segment-175830748773502782\_1580\_000\_1600\_000\_with\_camera\_labels.tfrecord} \\
& \texttt{segment-14561791273891593514\_2558\_030\_2578\_030\_with\_camera\_labels.tfrecord} \\
\bottomrule
\end{tabular}
\caption{Selected 20 evaluation sequences with official file names from the WOD dataset.}
\label{tab:wod_seq}
\end{table}

\begin{table}[h]
\centering
\small
\renewcommand{\arraystretch}{1.05}
\setlength{\tabcolsep}{8pt} 
\begin{tabular}{c c} 
\toprule
\textbf{Dataset} & \textbf{Sequence Indices} \\
\midrule
\multirow{2}{*}{NuScenes}
& 022 \quad 025 \quad 031 \quad 034 \quad 049 \quad 053 \quad 084 \quad 086 \quad 089 \quad 096 \\
& 210 \quad 323 \quad 350 \quad 382 \quad 403 \quad 410 \quad 430 \quad 534 \quad 570 \quad 640 \\
\bottomrule
\end{tabular}
\caption{Selected 20 evaluation sequences from NuScenes. Each row contains 10 sequence indices.}
\label{tab:nusc_seq}
\end{table}



\section{Discussion on the Availability of Feed-forward 3DGS}

\begin{figure}[h]
  
  \centering
  \includegraphics[width=0.9\linewidth]{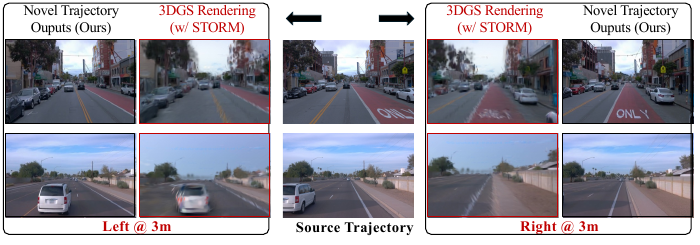}
  \caption{The effect of using the feed-forward 3dgs method during inference.}
  \label{supp_storm}
\end{figure}

To further evaluate the applicability of feed-forward 3DGS methods during inference, we use STORM~\cite{storm} to generate 20-frame 3DGS segments, which are then rolled out and concatenated to form the 121-frame 3DGS condition for our model inference. As shown in Fig.~\ref{supp_storm}, our method remains effective even when conditioned on these lower-quality renderings. This demonstrates that our framework is compatible with feed-forward 3DGS methods at inference time, highlighting its robustness and scalability.

\section{Temporal Consistency Error}

Inspired by the Photometric Consistency metric in WorldScore~\cite{worldscore}, we evaluate the inter-frame temporal stability of generated videos by computing the Temporal Consistency Error (TCE). This metric captures the Average End-Point Error across the entire video sequence. For each pair of consecutive frames $(I_t, I_{t+1})$, we extract a set of $N$ points $\mathbf{x}_t$ from a center-cropped region. This sampling strategy prioritizes central dynamic subjects while mitigating ``out-of-bounds" artifacts caused by camera motion.

We employ the off-the-shelf optical flow estimator SEA-RAFT~\cite{SEA-RAFT} to compute the forward optical flow $\phi_{f}$, yielding the predicted coordinates in the subsequent frame: $\hat{\mathbf{x}}_{t+1} = \mathbf{x}_t + \phi_{f}(\mathbf{x}_t).$
To verify tracking reliability, we perform a forward-backward consistency check by applying the backward optical flow $\phi_{b}$ to project the points back to the initial frame: $\mathbf{x}'_t = \hat{\mathbf{x}}_{t+1} + \phi_{b}(\hat{\mathbf{x}}_{t+1}).$ 
The consistency error $E(I_t, I_{t+1})$ for a frame pair is defined as the mean Euclidean distance between the original and re-projected points. For a video consisting of $T$ frames, the final TCE is calculated as the arithmetic mean across all consecutive pairs:
\begin{equation}
\mathcal{L}_{\text{TCE}} = \frac{1}{T-1} \sum_{t=1}^{T-1} E(I_t, I_{t+1}).
\end{equation}

\section{Comparisons of Different Training Stages}
\begin{figure}[h]
\setlength{\abovecaptionskip}{5pt}
\centering{\includegraphics[width=0.8\linewidth]{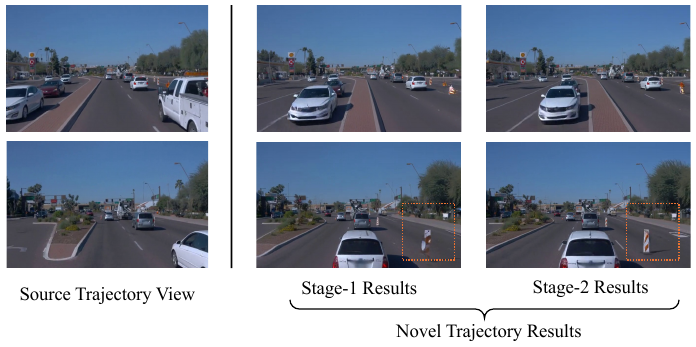}}
\caption{Comparison of results across different training stages. The first and second rows visualize synthesis results with a lateral offset of -4 m (left) and +4 m (right), respectively.}
\label{fig:supp_stage}
\end{figure}

Our framework adopts a progressive two-stage training paradigm to achieve precise camera control and consistent content generation. We compare the outputs of these two stages in Fig.~\ref{fig:supp_stage}. As illustrated, the results from Stage 1 occasionally exhibit imprecise viewpoint control (e.g., the first row), as this stage relies solely on relative camera poses for conditioning. In contrast, incorporating 3DGS renderings in Stage 2 not only enhances pose accuracy but also significantly improves the geometric consistency of the synthesized content, as highlighted by the orange dashed boxes in the second row.

\section{More Comparisons}
StreetCrafter is trained on Vista~\cite{vista}, while FreeVS is built upon Stable Video Diffusion (SVD)~\cite{svd}. Their frame lengths are summarized in Table~\ref{tab:frame_length}. We use the officially released rollout inference code to generate full-length sequences on WOD. However, because both models generate only short temporal windows, the rollout procedure often introduces background inconsistencies and noticeable temporal jitter. For a clearer assessment, please refer to the video results in our webpage.


\begin{table}[t]
\centering
\small
\renewcommand{\arraystretch}{1.2}
\setlength{\tabcolsep}{17.4pt} 
\begin{tabular}{l c c}
\toprule
\textbf{Method} & \textbf{Base Model} & \textbf{Frame Length} \\
\midrule
FreeVS & SVD~\cite{svd} & 8 \\
StreetCrafter & Vista~\cite{vista} & 25 \\
Ours & Wan2.1~\cite{wan} & 121 \\
\bottomrule
\end{tabular}
\caption{Frame length comparison of different methods.}
\label{tab:frame_length}
\end{table}

\section{More Qualitative Results}
\begin{figure*}[h]
\centering{\includegraphics[width=\linewidth]{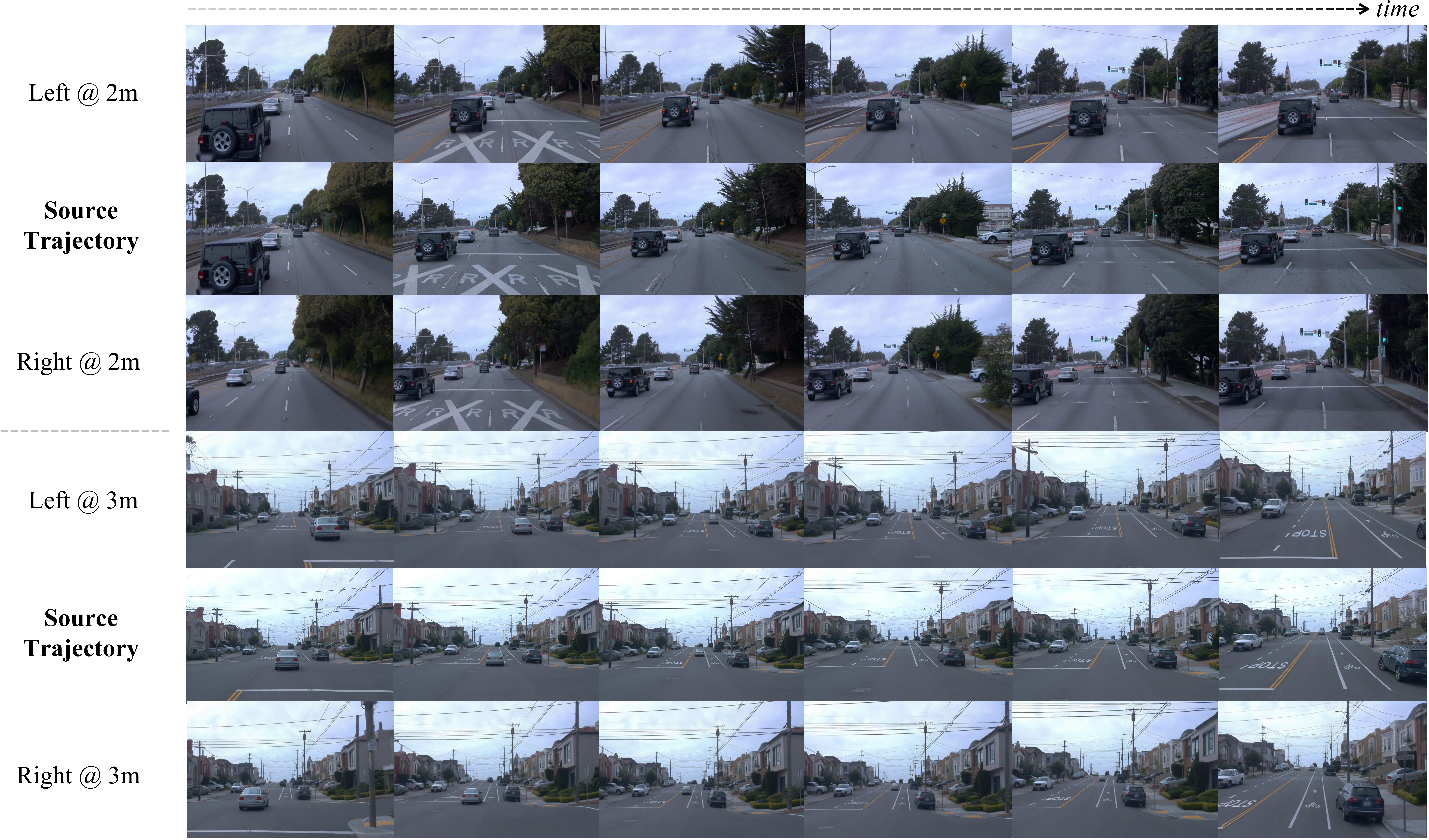}}
\caption{Qualitative results of our method on multiple novel-trajectory generations.}
\label{fig:demo1}
\vspace{-1.3em}
\end{figure*}

\begin{figure*}[h]
\centering{\includegraphics[width=\linewidth]{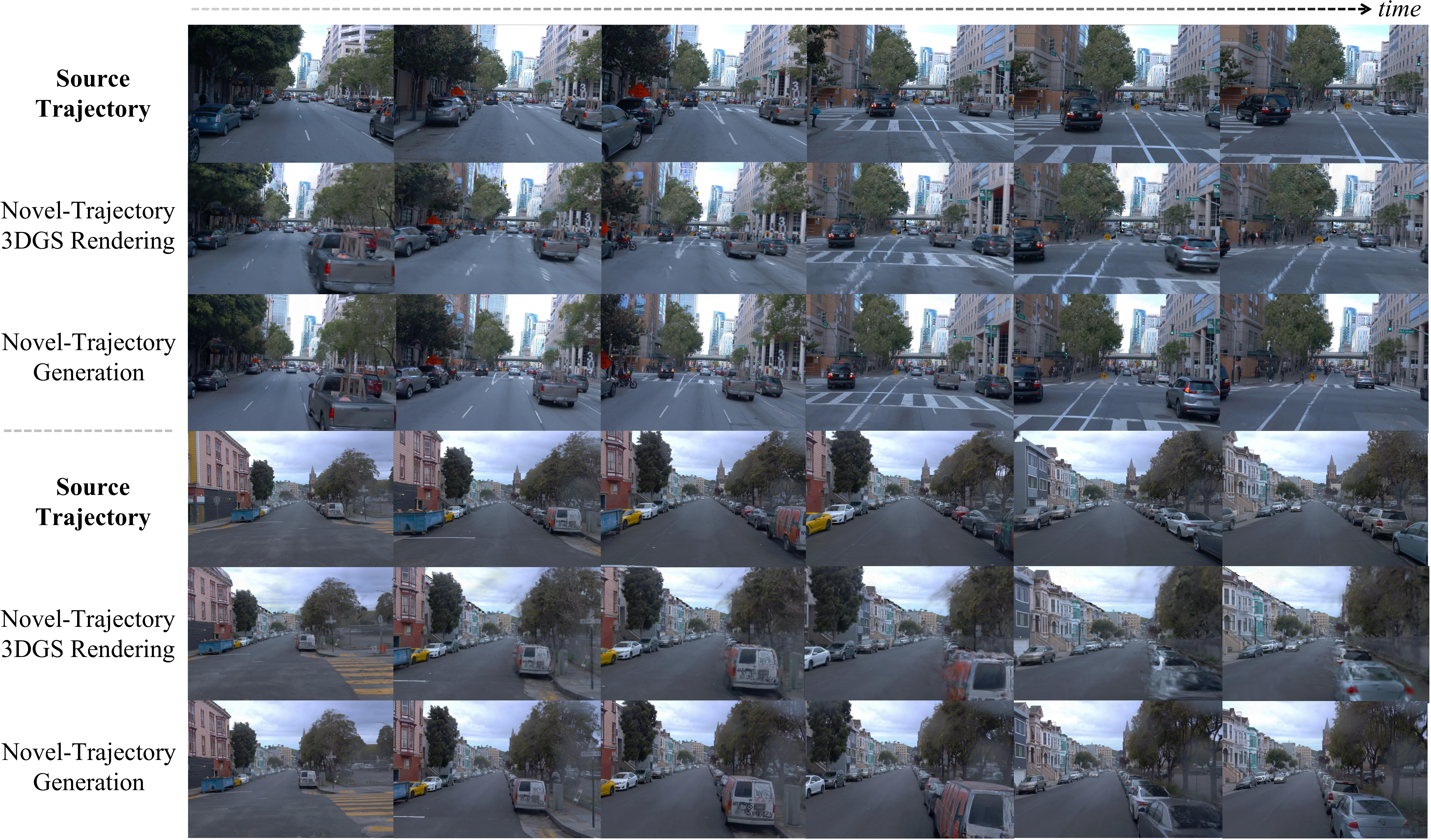}}
\caption{Qualitative novel-trajectory results and 3DGS rendering conditions, with the two scenes shifted right by 3 m and 4 m.}
\label{fig:demo2}
\vspace{-1.3em}
\end{figure*}

More qualitative results are shown in Fig.~\ref{fig:demo1} and Fig.~\ref{fig:demo2}, where our method maintains geometric consistency and visual fidelity while accurately generating novel trajectories. For clearer and more intuitive comparisons, we recommend viewing the video results in our webpage.

\end{document}